\documentclass{article} 
\usepackage[preprint]{colm2025_conference}

\usepackage{microtype}
\usepackage{hyperref}
\usepackage{url}
\usepackage{booktabs}
\usepackage{graphicx}
\usepackage{wrapfig}

\usepackage{hyperref}

\usepackage{algpseudocode}
\usepackage{algorithm}
\usepackage{array}
\usepackage{longtable}
\usepackage{wrapfig}
\usepackage{adjustbox}
\usepackage{multirow}
\usepackage{subcaption}
\usepackage{amsmath}

\usepackage{lineno}

\definecolor{darkblue}{rgb}{0, 0, 0.5}
\hypersetup{colorlinks=true, citecolor=darkblue, linkcolor=darkblue, urlcolor=darkblue}

\title{Stealing Creator's Workflow: A Creator-Inspired Agentic Framework with Iterative Feedback Loop for Improved Scientific Short-form Generation}

\author{Jong Inn Park, Maanas Taneja, Qianwen Wang, Dongyeop Kang\\
University of Minnesota\\
\texttt{\{park2838,tanej017,qianwen,dongyeop\}@umn.edu} 
}

\begin{document}

\ifcolmsubmission
\linenumbers
\fi

\maketitle

\begin{abstract}
Generating engaging, accurate short-form videos from scientific papers is challenging due to content complexity and the gap between expert authors and readers. Existing end-to-end methods often suffer from factual inaccuracies and visual artifacts, limiting their utility for scientific dissemination. To address these issues, we propose \textit{SciTalk}, a novel multi-LLM agentic framework, grounding videos in various sources, such as text, figures, visual styles, and avatars. Inspired by content creators' workflows, \textit{SciTalk} uses specialized agents for content summarization, visual scene planning, and text and layout editing, and incorporates an iterative feedback mechanism where video agents simulate user roles to give feedback on generated videos from previous iterations and refine generation prompts. Experimental evaluations show that \textit{SciTalk} outperforms simple prompting methods in generating scientifically accurate and engaging content over the refined loop of video generation. Although preliminary results are still not yet matching human creators' quality, our framework provides valuable insights into the challenges and benefits of feedback-driven video generation.
Our code, data, and generated videos will be publicly available.\footnote{\url{https://minnesotanlp.github.io/scitalk-project-page/}}
\end{abstract}

\section{Introduction}
With the rise of short-form video platforms like Shorts, TikTok, and Reels, the public increasingly encounters scientific and technological content through brief, engaging short-form videos. Currently, most of these videos are produced by expert human creators who combine strong scientific backgrounds with storytelling and video editing skills to craft compelling content. In contrast, automatic generation of such videos, particularly when they involve complex scientific topics, remains a significant challenge.

Recent diffusion-based video generation models \citep{10.5555/3495724.3496298} have shown potential by generating full end-to-end videos from various inputs such as text, images, or existing clips, relying on the models’ generative capabilities. However, these fully generated videos often struggle with factual accuracy and tend to produce visual artifacts. Even with carefully engineered prompts and agentic pipeline, these generative models fall short generating precise images for scientific knowledge due to its permissible creative contexts or providing content in the format of short reports or images.

To enable the generation of short-form scientific videos that remain faithfully grounded in the source material, we draw inspiration from the workflows of human content creators. Empirical studies show that short-form video creators typically follow an iterative, multi-stage process, including planning, production, and editing—to craft content that aligns with their intended message \citep{Klug_2020, 10.1145/3544548.3581386, inproceedings, ghosh2023establishing}. Creators often iterate on their videos, modifying clips, layering effects, adding text, and incorporating images to ensure the final product meets their communicative goals.

Building on these insights, we introduce \textit{SciTalk}, a fully automatic, agentic, and creator-inspired video generation framework tailored to the unique demands of scientific short-form content. 
\textit{SciTalk} orchestrates a suite of specialized LLM-based agents that collaborate across four major stages: Preprocessing, Planning, Editing, and Feedback \& Evaluation. The final video is composed using a video editing library, not generative models, ensuring high visual fidelity and factual accuracy.
Importantly, \textit{SciTalk} integrates an iterative feedback loop inspired by human creators' iterative review processes. After a video is composed, vision-language model based \textit{Feedback Agents} evaluate each sub-scene using both qualitative and quantitative metrics. \textit{Reflection Agents} then incorporate feedback into revised prompts, progressively refining video quality and aligning it more closely with the intended scientific narrative.

Our experiments suggest that our workflow produces videos that are more engaging and informative than those generated by a simple single-agent baseline. However, the generated videos still fall short of creator-produced content in terms of coherence and polish, and we observe that errors may accumulate over iterations. Despite these limitations, our work serves as a preliminary exploration into automatic, agent-driven scientific video generation. To the best of our knowledge, this is the first study to tackle this task, and our findings offer early insights into the challenges in this emerging area. We envision future work aimed at optimizing the workflow and improving system robustness.

Our primary contributions include:
\begin{itemize}
    \item We propose a creator-inspired agentic workflow explicitly designed for grounding short-form videos in scientific materials. The collaboration among multiple specialized agents enables controllable, accurate, and high-quality video creation.
    \item We introduce an iterative feedback mechanism, allowing the system to incrementally improve video quality and alignment with the user engagement through the video feedback model.
    \item We conduct a comprehensive evaluation across multiple metrics and sources, providing early empirical insights into the challenges of agentic video generation, such as feedback misalignment and visual clutter, while highlighting areas where structured, iterative generation shows promise.
\end{itemize}

\section{Related Work}

\paragraph{Automatic Short-form Video Generations}
Recent video generation task uses diffusion models to create realistic and consistent video sequences. The first video diffusion model~\citep{NEURIPS2022_39235c56} showed how iterative noise reduction could synthesize smooth transitions, while MCVD~\citep{voleti2022MCVD} introduced a masked conditional framework to handle various video tasks, including future prediction and interpolation. Furthermore, GAN and VAE models~\citep{digan} have been explored for long-form video generation. Make-A-Video~\citep{singer2023makeavideo} enabled text-to-video generation without paired text-video datasets, instead learning from text-image pairs and unlabeled videos.

However, keeping consistency still remains a challenge; To tackle this issue, ARTDiff~\citep{lu2024improve} introduced correlated noise across frames to ensure smoother motion, while ConsistI2V~\citep{ren2024consistiv} leveraged spatiotemporal attention to preserve consistency in image-to-video transformations. Meanwhile, OpenAI's Sora~\citep{videoworldsimulators2024} unveils a high-resolution video generation model. Furthermore, \citet{10.1145/3477314.3507141} explores video editing techniques, resulting in a new video with various types of shots and camera movements. In addition, \citet{8354295} investigates summarizing of a video by clipping highlighted segments within the original context. In summary, research in video generation generates synthetic or source-based videos.

\paragraph{Multi-agent Workflow}
Multi-agent systems have emerged to decompose the generation process into specialized roles, often inspired by human collaboration. Kubrick~\citep{he2024kubrickmultimodalagentcollaborations}, GenMac~\citep{huang2024genmaccompositionaltexttovideogeneration}, Mora~\citep{yuan2024moraenablinggeneralistvideo}, VideoAgent~\citep{soni2025videoagent}, and FilmAgent~\citep{xu2025filmagent} adopt collaborative, LLM-powered agents with coordinator components to guide the workflow and inject feedback through iterative loops. Other works like AesopAgent~\citep{wang2024aesopagentagentdrivenevolutionarystorytovideo} and StoryAgent~\citep{hu2024storyagentcustomizedstorytellingvideo} focus on converting stories into videos using similar agentic pipelines. SPAgent~\citep{tu2024spagentadaptivetaskdecomposition} leverages agents powered by various diffusion models to generate scenes from diverse inputs (e.g., text, image, video). In our study, we similarly assign agents to specific roles and enhance the workflow with a feedback-driven refinement loop.

\paragraph{Visual Dissemination of Scientific Knowledge}
The intrinsic complexity of scientific knowledge makes reading, reviewing, and writing scholarly documents a cognitively demanding task \citep{kang-etal-2018-dataset,wang2025scholawrite,head2021augmenting}. 
To enhance the dissemination of that knowledge, researchers have begun converting scientific papers into structured visual summaries that deliver key information more efficiently and succinctly.
Doc2PPT~\citep{Fu2021DOC2PPTAP} converts academic papers into presentation slides, integrating text summaries and actual figures to highlight key findings. Similarly, \citet{kumar-etal-2024-longform} built a dataset of 3,000 papers paired with conference presentations, training a model to generate science blogs that merge textual explanations with visual elements. Furthermore, Pub2Vid~\citep{10.1145/3573381.3596157} helps researchers in generating video abstracts, suggesting key sentences, figures, and simple animations to streamline the video creation process. To get multi-modal models grounded on real scientific knowledge, datasets such as MultiModal ArXiv~\citep{li-etal-2024-multimodal-arxiv} and WikiWeb2M~\citep{burns2023wiki}, provide resources for training models that align textual content with relevant visuals. Aligned with efforts in grounding visual components to actual materials, our work tries to preserve paper sources.

\section{Methodology}

\paragraph{Creator's workflow}
Short-form videos have become an increasingly popular medium for communicating scientific knowledge across platforms like TikTok, Instagram, X (formerly Twitter), and YouTube. \footnote{%
\label{fn:video_examples}
\begin{tabular}[t]{@{}l@{}}
\url{https://x.com/victorialslocum/status/1897612417237983580/video/1} \\
\url{https://x.com/victorialslocum/status/1856334367724859410/video/1} \\
\url{https://www.youtube.com/shorts/lYSm55rgY8M}
\end{tabular}
}
Some creators specifically focus on summarizing and discussing research papers, often by showing the PDF, narrating key points, and using voiceovers. These videos, while sometimes longer than typical platform averages due to the complexity of the content, often rely heavily on raw materials from the paper—figures, screenshots, or text snippets—which helps ground the video in factual, source-based information. This grounding is essential to maintain scientific integrity and fulfill the core goal of clear, trustworthy communication. Prior studies indicate that successful video production involves a cycle of planning, production, and iterative refinement until the final product aligns with the creator’s process~\citep{Klug_2020, 10.1145/3544548.3581386, inproceedings, ghosh2023establishing}.

\begin{figure}[t]
    \centering
    \includegraphics[width=\linewidth]{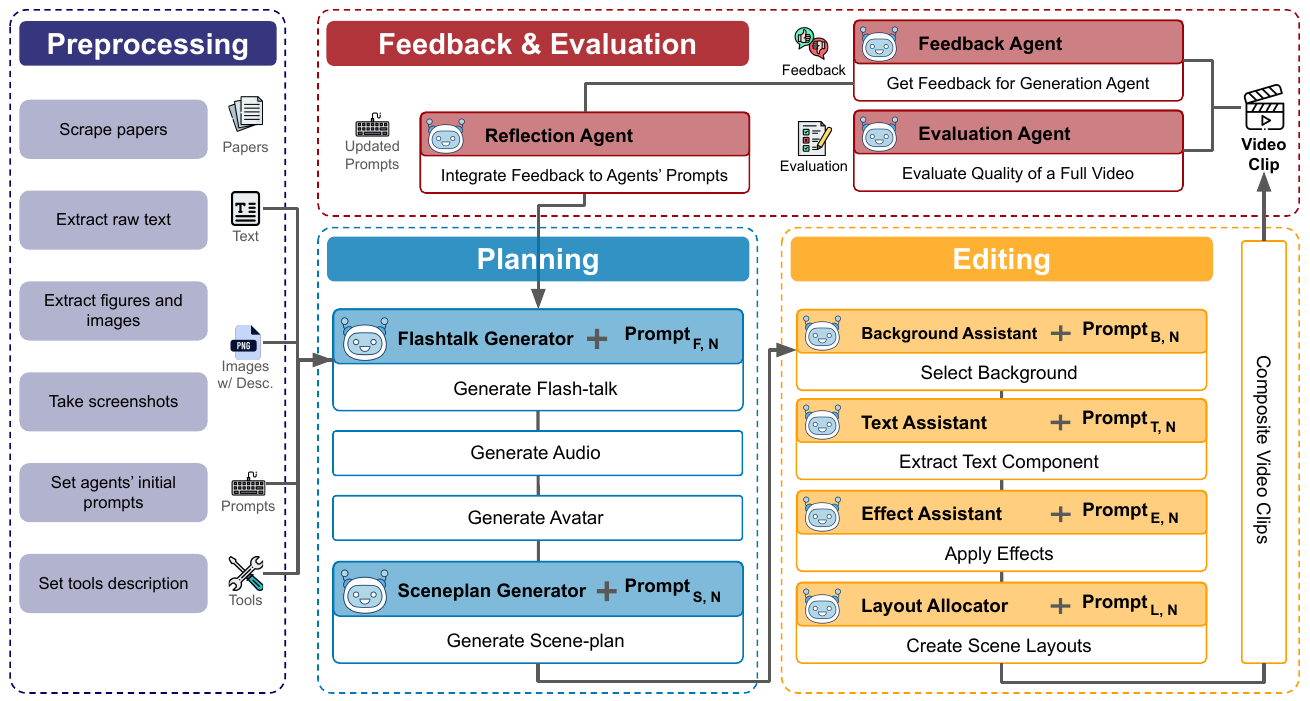}
    \caption{Conceptual overview of the multi-agent video generation pipeline. The pipeline comprises four distinct stages: (1) Preprocessing Stage, where papers are scraped to extract raw text, figures, images, and screenshots; (2) Planning Stage, involving specialized assistants that generate detailed outputs (audio, flash-talk, scene-plan, avatar, background, etc.) to composite a video clip; (3) Editing Stage, which integrates visual effects, image layouts, and text components; and (4) Feedback \& Evaluation Stage, assessing video components, reflecting the improvements in feedback to next prompts, and evaluating overall quality of the final video output. Notation: $\text{Prompt}_{i,j}$ denotes the prompt used by agent $i$ during the $j$-th iteration, where $i \in \{F, S, B, T, E, L\}$ corresponds to each generation agent.
    }
    \label{fig:pipeline}
\end{figure}

\subsection{Multi-Agent Workflow} 
Inspired by this human thought process, we design our framework with four stages: preprocessing, planning, editing, and feedback and evaluation stages. The pipeline (See Figure \ref{fig:pipeline}) consists of six LLM-based agents:\textit{Flashtalk Generator}, \textit{Sceneplan Generator}, \textit{Background Assistant}, \textit{Text Assistant}, \textit{Effect Assistant}, and \textit{Layout Allocator}. 
Each agent receives structured input materials and returns formatted outputs, enabling modular, coordinated generation.

\textbf{Preprocessing Stage.} This stage prepares all relevant materials from the input paper for downstream agents. The pipeline begins by scraping papers from sources like arXiv and extracting key content—text (from HTML or PDF), figures, tables, and screenshots (e.g., the first page of the paper). These assets are stored in a structured format and reused in later stages to ensure the final video remains grounded in the original scientific source.

\textbf{Planning Stage.} Two main agents operate in this stage: First, \textit{Flashtalk Generator} creates an overarching script and structure for the video, modeled after the flash-talk format used at academic conferences\footnote{A \textit{flash-talk} is a 1–3 minute presentation designed to quickly engage audiences at conferences e.g., \url{https://www.youtube.com/watch?v=2MLEq6-zk1A} \citep{researchamerica2020flashtalks}. }. This initial script includes four structured components: (1) Aggressive Hook, (2) Brief Context, (3) Intriguing Teaser, and (4) Call to Action. Each section includes associated visual assets and serves both as a planning document and the voiceover script for the video. Additional submodules—like an \textit{Audio Assistant} and \textit{Avatar Assistant}—generate mp3/mp4 outputs for narration and avatar animations, respectively, which will be leveraged in Editing stage.

\textit{Sceneplan Generator} breaks down each section into 1–5 sub-scenes based on the script. Each sub-scene includes a textual description, timing metadata, and associated images. These scene descriptions provide guidance for editing agents, including prompts like "Start with a dramatic zoom-in on first\_page.png." This structured breakdown allows for fine-grained control and alignment between audio and visuals. The images assigned to the section are inherited and re-distributed to its sub-scenes with a duration attribute.

\textbf{Editing Stage.} This stage handles the generation and layout of all visual elements: \textit{Background Assistant} selects background images based on scene descriptions and available images. (Note: in current experiments, we use a fixed black background to focus on foreground content.) \textit{Text Assistant} generates subtitles and other on-screen text, specifying parameters such as font size, color, duration, and position, to ensure alignment between text overlays and the scene’s pacing.
Those parameters will be plugged in to the video clip creation module.  \textit{Layout Allocator} determines the visual layout of each sub-scene. This LLM-based agent receives a simplified HTML-style markup of the scene’s structure and returns positional data for each component. While the agent lacks visual reasoning abilities, the structured input enables consistent scene composition.

\textbf{Feedback \& Evaluation Stage.} 
Three feedback agents—each powered by a multi-modal LLM (MLLM)—review intermediate outputs: \textit{Flashtalk Feedback Agent}, \textit{Sceneplan Feedback Agent}, and \textit{Text Feedback Agent}. Because current MLLMs (e.g., LLaVa-NeXT-Video 34B~\citep{zhang2024llavanextvideo}) cannot process entire videos due to token limits, the feedback loop evaluates sub-scenes individually. Each feedback agent is guided by a set of role-specific metrics (see Appendix Table~\ref{tab:eval_rubric_feedback}) to ensure relevance—for example, the \textit{Flashtalk Feedback Agent} focuses only on narrative flow and audio script quality, not visuals.

To incorporate feedback, \textit{Reflection Agents} revise each generation agent's prompt based on relevant feedback. These agents filter out irrelevant suggestions and reformat useful ones for prompt refinement. Only the original prompt and current feedback are considered in this step to stay within token limits. Full algorithmic details are described in Appendix Algorithm~\ref{alg:feedback-loop}. Finally, a separate MLLM-based \textit{Evaluation Agent} provides an end-to-end assessment of the final video. It rates the overall output using human-aligned metrics and enables comparisons to videos created by human experts.

\subsection{Feedback Loop for Iterative Refinement}

\textbf{Feedback Generation \& Aggregation.} 
The goal of this loop is to enhance video quality and flow without directly altering agent behavior. Feedback is kept agent-specific: for instance, visual feedback is not passed to agents responsible for narrative flow. Each feedback agent uses a predefined metric set (detailed in Appendix Table~\ref{tab:eval_rubric_feedback}) to focus its evaluation. Since full-video input exceeds model limits, evaluation proceeds sub-scene by sub-scene in an iterative loop.

\textbf{Feedback Reflection.} \textit{Reflection Agents} selectively integrate relevant feedback into the next iteration’s prompts. For example, the \textit{Reflection Agent} for \textit{Sceneplan Generator} filters for feedback affecting pacing and coherence, ignoring irrelevant visual suggestions. This ensures each agent only acts on feedback within its domain, and avoids cascading inconsistencies across the pipeline.

\begin{figure}
    \centering
    \includegraphics[width=\linewidth]{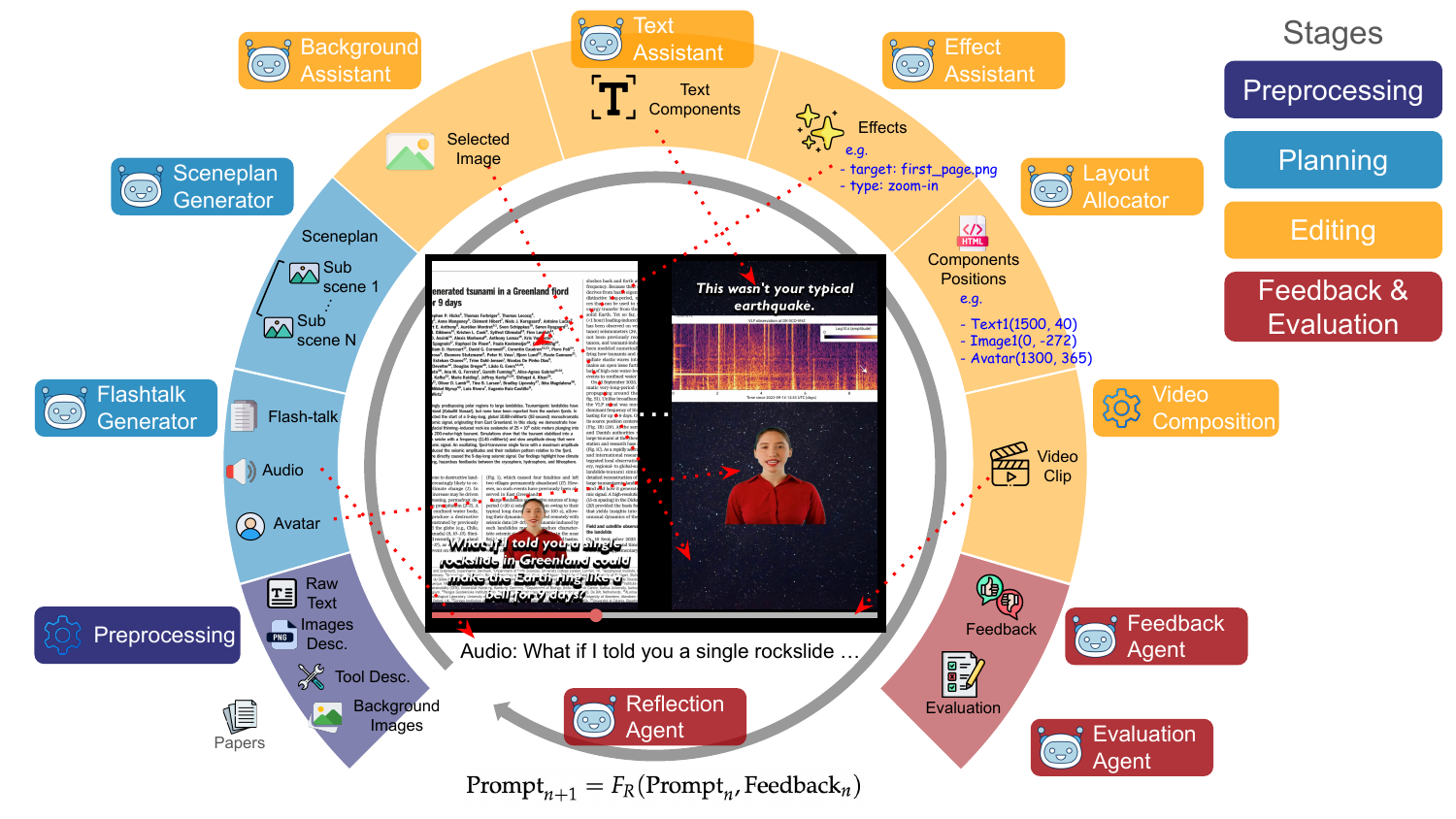}
    \caption{Detailed workflow on how generation agents contribute to scene composition. Agents operate sequentially across four stages, producing parameters that are passed to the video composition module for final assembly.}
    \label{fig:agents_involvement}
\end{figure}

\subsection{Sanity Check Before Final Composition}
To ensure the integrity of the final video output, \textit{SciTalk} includes a basic sanity-check mechanism applied just before video composition. Given the complexity of each agent's task, the generation process often produces hallucinated or unintended outputs. For example, despite explicit instructions to avoid text overlap or clutter, the \textit{Text Assistant} may still generate subtitles that visually interfere with other elements. To address these issues, we apply simple filtering rules to exclude clearly flawed outputs during the final assembly stage. However, in our main experiments, we deliberately disabled this mechanism to evaluate the full autonomy of the generation agents.

\subsection{Summary}
Figure~\ref{fig:agents_involvement} illustrates our video generation pipeline using \citet{doi:10.1126/science.adm9247} as an example. First, the \textit{Flashtalk Generator} produces a structured narrative (e.g., Vivid Hook) from preprocessed content (e.g., \texttt{first\_page.png}), with audio scripts and avatar animations generated via OpenAI~\footnote{\url{https://platform.openai.com/docs/models/whisper-1}} and Synthesia~\footnote{\url{https://www.synthesia.io/}} APIs. Next, the \textit{Sceneplan Generator} subdivides the narrative into sub-scenes with defined directions and timings. Editing agents (\textit{Background Assistant}, \textit{Text Assistant}, \textit{Effect Assistant}, and \textit{Layout Allocator}) prepare and composite visual elements using MoviePy~\citep{zulko2014}. Finally, \textit{Feedback} and \textit{Reflection Agents} iteratively refine the outputs, and an \textit{Evaluation Agent} assesses the final video using human-aligned criteria.

\section{Experiments}

\textbf{Seed Paper Selection.} To evaluate the effectiveness of the \textit{SciTalk} framework, we selected nine scientific papers across various domains. These papers were assigned abbreviated identifiers for reference:  \textsc{Rockslide}~\citep{doi:10.1126/science.adm9247}, \textsc{Qwen2.5}~\citep{qwen2.5}, \textsc{Knownet}~\citep{10670469}, \textsc{Tuning}~\citep{ahn-etal-2024-tuning}, \textsc{Threads}~\citep{kim-etal-2024-threads}, \textsc{Dynamic}~\citep{de-langis-etal-2024-dynamic}, \textsc{Query}~\citep{shorten2025queryingdatabasesfunctioncalling}, \textsc{Context}~\citep{morris2025contextual}, and \textsc{Speed}~\citep{Chen_2023_CVPR}. 
Among these, \textsc{Rockslide} represents non-computer science content, while \textsc{Qwen2.5} is an 18-page technical report.
Four of the selected papers—\textsc{Knownet}, \textsc{Tuning}, \textsc{Threads}, and \textsc{Dynamic}—were authored by members of our research group.
The remaining three papers were used for comparison with creator-generated content. For each paper, five iterations of videos were generated, resulting in a total of 45 ($9 \times 5$) videos.

\textbf{Models \& Implementation.} We use GPT-4o~\citep{openai2024gpt4ocard} as the backbone model for all generation agents. The \textit{Feedback} and \textit{Evaluation Agents} utilize LLaVA-NeXT-Video 34B~\citep{zhang2024llavanextvideo} to simulate human-like judgment and process full-length video contexts. Due to the long inference time resulting from the model’s size, we evaluated only a single metric per feedback agent in the experiments: \textit{Curiosity} for Flashtalk, \textit{Visual Relevance and Clarity} for Sceneplan, and \textit{Key Information Coverage} for Text (Table \ref{tab:eval_rubric_feedback}). All models operate on images at a resolution of $360\times640$, with the \textit{Flashtalk Feedback Agent} receiving 10 images per section, the \textit{Sceneplan} and \textit{Text Feedback Agents} 2 images per sub-scene, and the \textit{Evaluation Agent} 60 images per video. Human evaluation is conducted by our two internal annotators, and we simulate model diversity by sampling outputs with two temperatures (0.7 and 0.9); invalid outputs were excluded from aggregation.

\begin{small}
\begin{longtable}{p{1.2cm}|p{3.6cm}|p{8cm}}
\toprule
\textbf{Section} & \textbf{Metric} & \textbf{Description} \\ 
\midrule
\endhead
Content Accu- & Scientific Integrity (\textbf{SI}) & Measures how well the video adheres to the core findings of the paper without distortion \\
racy & Key Concept Coverage (\textbf{KCC}) & Assesses whether the video includes the main hypotheses, methods, results, and conclusions. \\
 
\cmidrule{1-3}

Clarity & Logical Flow (\textbf{LF}) & Checks if the summary follows a logical sequence (e.g., Background → Methods → Results → Conclusion). \\

 & Comprehensibility (\textbf{C}) & Evaluates whether the explanation is understandable for a general audience while maintaining accuracy. \\
\cmidrule{1-3}

Visual \& Audio  & Scene Readability (\textbf{SR}) & Ensures that any text on screen is clear, not too dense, and easy to read. \\

Sync & Audio-Visual Alignment (\textbf{AVA}) & Evaluates if spoken content matches the on-screen visuals at the right moment. \\
\cmidrule{1-3}

Engag- ement & Attention Retention (\textbf{AR}) & Predicts how well the video keeps viewers engaged throughout. \\

 & Pacing & Evaluates whether the video moves too fast or too slow for comprehension. \\

 & Call-to-Action Effectiveness (\textbf{CTA}) & Checks if viewers are encouraged to read the full paper or visit related resources. \\

 & Highlight Emphasis (\textbf{HE}) & Measures if key findings are properly emphasized through animation, captions, or repetition. \\
\bottomrule

\caption{Evaluation rubrics for human and \textit{Evaluation Agent}.}
\label{tab:eval_rubric_hm}
\end{longtable}
\end{small}

\textbf{Evaluation Metrics.} Since there is no existing benchmark directly comparable to our framework, we designed a custom evaluation rubric capturing four key aspects: content accuracy, clarity, visual-audio synchronization, and engagement. Each of these categories includes fine-grained sub-metrics, as detailed in Table~\ref{tab:eval_rubric_hm}.
These metrics will be used for the overall video qualification assessment by humans or off-the-shelf multimodal models.

\textbf{Research Questions.}
We address the following questions:
\begin{itemize}
    \item RQ1. How does the iterative feedback refinement process impact the generated scientific videos? (\ref{sec:progressive})
    \item RQ2. Does the specialized multi-agentic pipeline improve the video quality against single-agent or other generation-based approaches? (\ref{sec:agentic})
    \item RQ3. To which extent do generated full videos meet quality evaluations in terms of content alignment, comprehensibility, visual coherence, and engagement ? Do human and vision-language model similarly assess video quality improvements over iterations? (\ref{sec:full})
    \item RQ4. How do videos produced by the framework compare to videos created by human experts? (\ref{sec:human-vs-scitalk})
    \item RQ5. What specific qualitative strengths and limitations emerge from analyzing representative examples of videos produced by our framework? (\ref{sec:qualitative})
\end{itemize}

\subsection{Progressive Effect of Iterative Refinement}\label{sec:progressive}
\begin{figure}[htbp]
    \centering
    \begin{subfigure}[b]{0.48\linewidth}
        \centering
        \includegraphics[height=4cm]{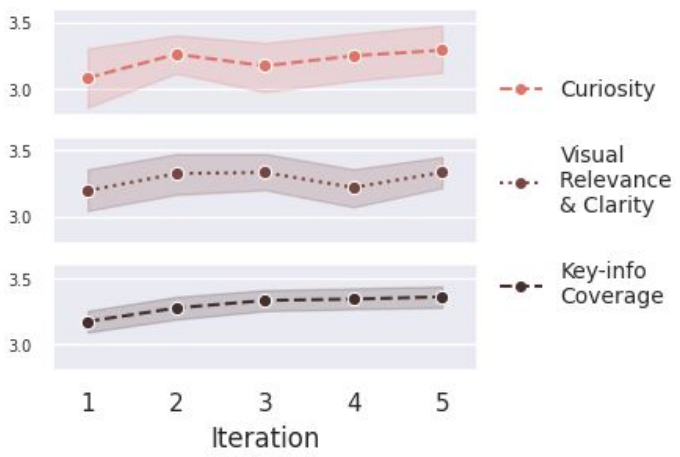}
        \caption{Averaged feedback metrics scores by iteration.}
        \label{fig:feedback_by_iter}
    \end{subfigure}
    \hfill
    \begin{subfigure}[b]{0.48\linewidth}
        \centering
        \includegraphics[width=\linewidth]{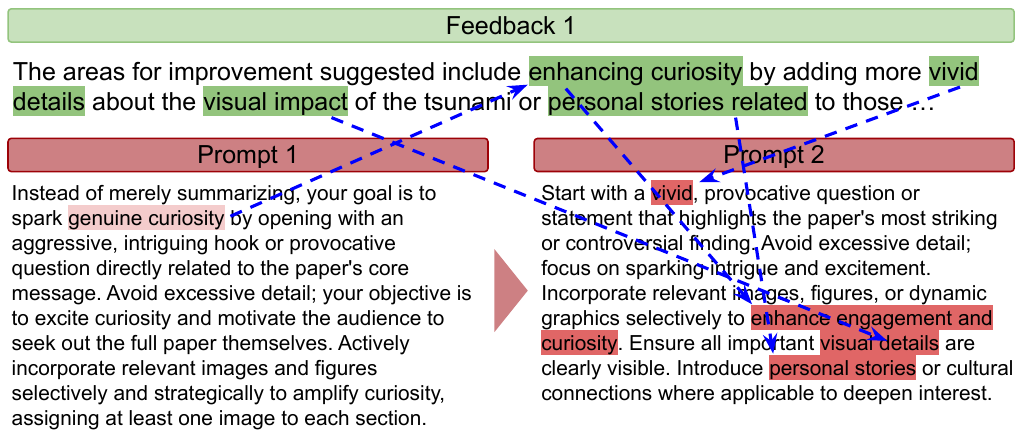}
        \caption{Comparison between 1st and 2nd iteration prompt for \textsc{Rockslide} paper, demonstrating how feedback is  integrated into subsequent prompts.}
        \label{fig:prompt_diff}
    \end{subfigure}
    \caption{Improvements on feedback metrics and prompts.}
    \label{fig:feedback_scores_prompts}
\end{figure}

To assess the effectiveness of iterative feedback, we examined how scores assigned by feedback agents evolved across iterations. Sub-scenes were evaluated independently, using both the generation agent's outputs and the associated scene data. Figure~\ref{fig:feedback_by_iter} reveals slight but consistent improvements in agent-assigned scores across iterations, despite some fluctuations. Although the increase is modest, it indicates that iterative feedback is progressively enhancing output quality. The subtle upward trend observed underscores the utility of the feedback mechanism, yet it also highlights potential areas for refinement in feedback precision or model responsiveness. In addition to measuring score trends, we tracked how feedback was integrated into prompt revisions across iterations. As illustrated in Figure~\ref{fig:prompt_diff}, we analyzed whether feedback from iteration $i$ was reflected in the prompt for iteration $i+1$, offering insight into the prompt update mechanism and the quality of our feedback incorporation loop.

\subsection{Effectiveness of Multi-Agentic Pipeline}\label{sec:agentic}
We further evaluated the impact of our specialized multi-agent pipeline in \textit{SciTalk} by comparing it with a single-agent baseline. In this baseline, a single prompt was used to generate the entire video in one pass, without intermediate agents or modular planning. Since this approach lacks the structured handoffs required by downstream modules, the prompt was simplified to produce direct end-to-end outputs. Results (Figure~\ref{fig:hm_eval_averaged}) demonstrate that the multi-agent system improves compositional quality, coherence, and alignment with scientific messaging.

\subsection{Sub-Optimal Gains in Quality of Generated Full Videos}\label{sec:full}

\begin{figure}
    \centering
    \noindent\includegraphics[width=\linewidth]{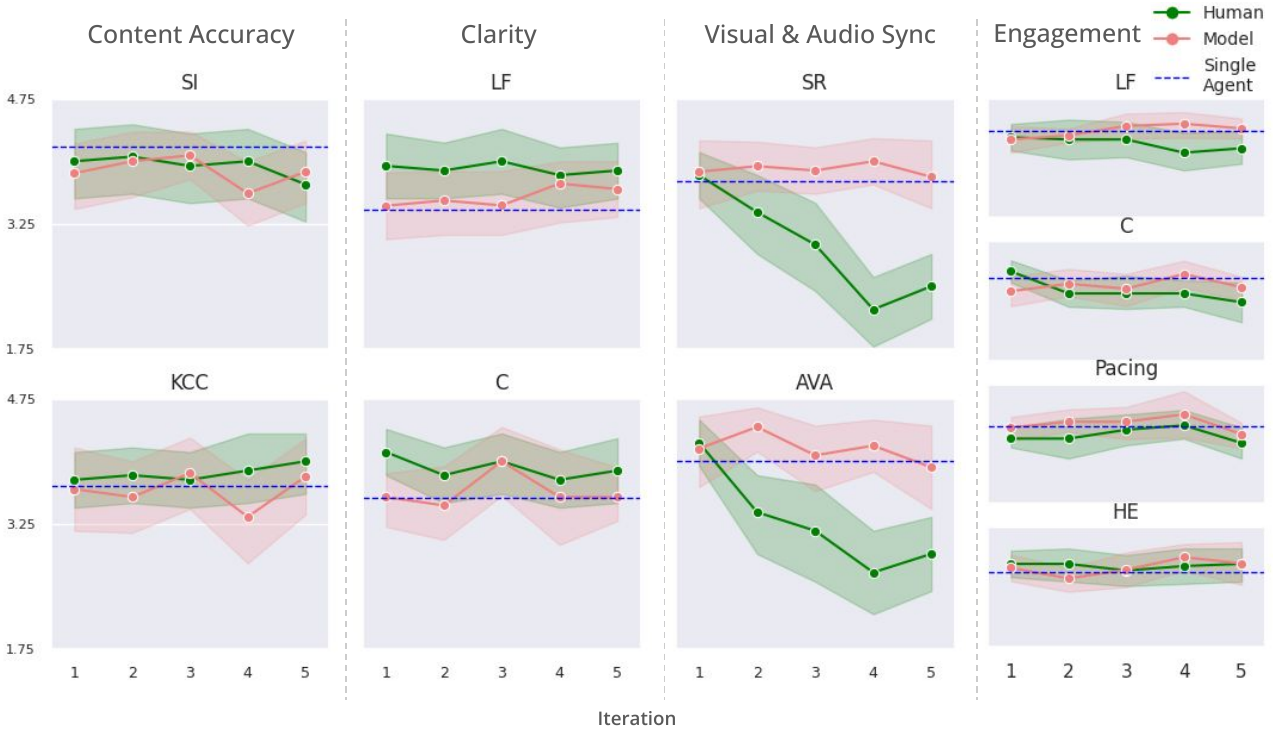}
    \caption{Average evaluation scores across iterations for both human and model evaluations. Shaded regions represent 95\% confidence intervals. All score axes are standardized to a range between 1.75 and 4.75 for consistency across metrics; higher scores indicate better performance. The blue dashed line represents the model-only average score from a single-agent baseline.}
    \label{fig:hm_eval_averaged}
\end{figure}

We analyzed evaluation scores across all five iterations of each video to assess quality progression. As shown in Figure~\ref{fig:hm_eval_averaged}, model scores in the engagement dimension generally improved over time, peaking at the 4th iteration. In addition, model scores in content accuracy peaked at the 3rd iteration, dipped slightly at the 4th iteration, and exhibited modest recovery by the 5th iteration. However, scores related to visual and audio synchronization fluctuated and showed a slight downward trend.

In contrast, human evaluations showed minor fluctuations across iterations, with significant declines observed especially in visual-audio synchronization metrics (SR: 3.8 $\rightarrow$ 2.2; AVA: 4.2 $\rightarrow$ 2.7). This discrepancy likely stems from model evaluations lacking precise timestamp alignments and overlooking visual clutter introduced by longer scripts and denser subtitles. These results highlight a clear gap between model-driven refinements and human perception, underscoring the need for more human-aligned evaluation strategies.

\subsection{Comparison against Creators' Videos}\label{sec:human-vs-scitalk}
\begin{figure}[ht]
    \centering
    \noindent\includegraphics[width=\linewidth]{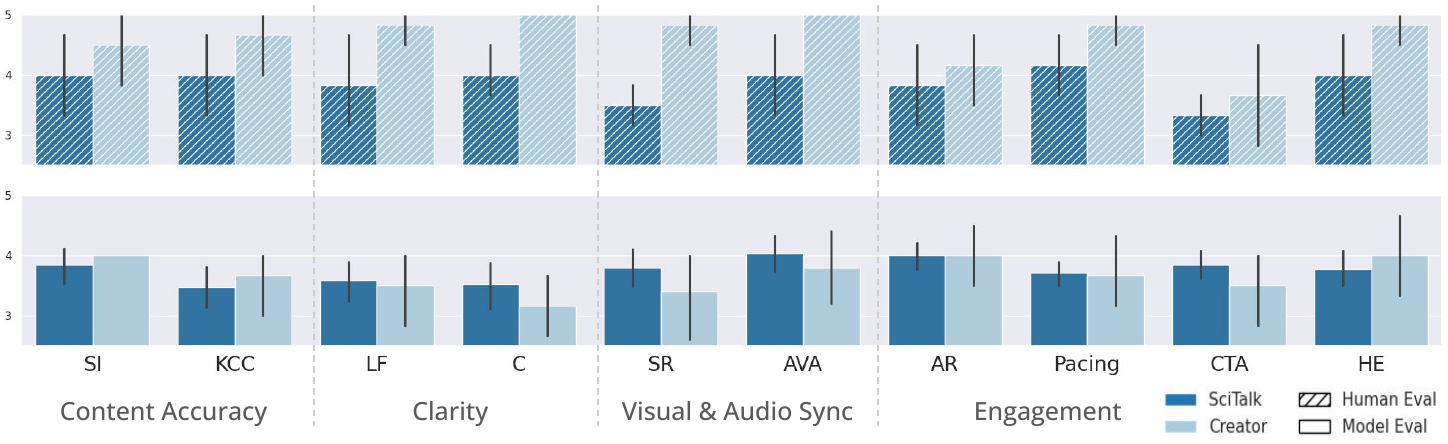}
    \caption{Comparison of evaluation scores across three papers (\textsc{Query}, \textsc{Context}, and \textsc{Speed}) between \textit{SciTalk}-generated and human Creator videos. \textbf{Top:} direct comparison of the 1st \textit{SciTalk} iteration against Creator videos. \textbf{Bottom:} comparison of the mean score across the five iterations against Creator videos. Black error bars denote 95\% confidence intervals.}
    \label{fig:hm_eval_creator}
\end{figure}

To understand how \textit{SciTalk}-generated videos compare to those made by human creators, we conducted a focused comparison using three papers: \textsc{Query}, \textsc{Context}, and \textsc{Speed} (See footnote~\ref{fn:video_examples}). The evaluations utilized two distinct settings. In the human evaluation (Figure~\ref{fig:hm_eval_creator}, top), we specifically compared the 1st iteration of \textit{SciTalk} videos against human creators' videos, as the 1st iteration was generally rated highest by human evaluators. As expected, human evaluators generally preferred creator-made videos, likely reflecting creators' expertise and experience in content production. Conversely, in the model evaluation (Figure~\ref{fig:hm_eval_creator}, bottom), we compared the average performance across the five iterations of \textit{SciTalk} against human-created videos. Interestingly, the model evaluation favored \textit{SciTalk}-generated videos in several metrics. Specifically, \textit{SciTalk} videos outperformed creator-made videos in clarity-related metrics (LF and C), highlighting its effectiveness in organizing content structure clearly. Additionally, \textit{SciTalk}-generated videos scored slightly higher in visual and audio synchronization (SR and AVA), indicating strong alignment between visual elements and narration. While these results showcase the feasibility of the \textit{SciTalk} framework from the model evaluation, they also reveal that it has yet to meet the high standards set by human creators.

\begin{figure}
    \centering
    \includegraphics[width=\linewidth]{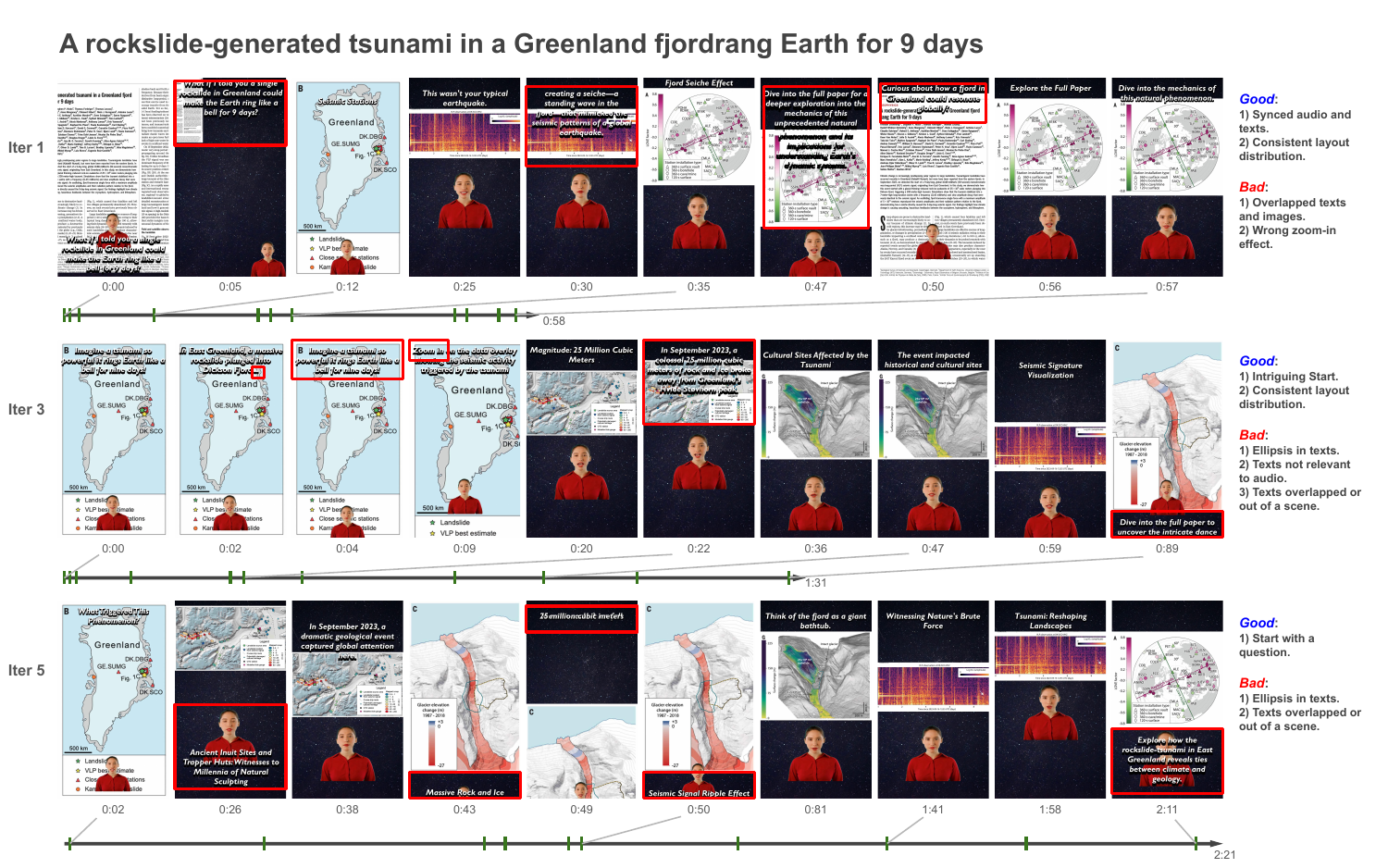}
    \caption{Qualitative analysis of iterative refinement effects on videos for \textsc{Rockslide}. The comparison across Iterations 1, 3, and 5 highlights diverse video structures. Persistent issues, including overlapped texts, irrelevant content alignment, and improper zoom effects, illustrate areas for continued refinement.}
    \label{fig:qual_rock}
\end{figure}

\subsection{Qualitative Examples}\label{sec:qualitative}

Finally, we conducted a qualitative analysis of how \textit{SciTalk} compositions evolve through iterations. As shown in Figure~\ref{fig:qual_rock}, early iterations often featured layout misalignments and poorly distributed text. Despite feedback, the \textit{Text Assistant} continued to occasionally generate ellipses or overlapping content. However, the 
\textit{Flashtalk Generator} showed notable improvements, restructuring sections based on prior feedback. This restructuring also influenced the \textit{Sceneplan Generator}, which adjusted the number and composition of sub-scenes in later iterations, often increasing scene richness and video duration. These examples illustrate both the benefits and challenges of feedback-driven refinement.

\section{Discussion and Limitation}
Our framework, \textit{SciTalk}, simulates human-like workflows using modular agents and iterative feedback loops. While this structured pipeline refines content, the feedback loop can increase visual complexity and degrade clarity, and model-generated feedback often diverges from human evaluations—especially in subjective metrics like audio-visual alignment and engagement. Additionally, autonomous prompt regeneration may propagate small changes, leading to instability.

Future work should focus on three areas: (1) learning structured video-generation patterns from human examples to streamline prompt optimization, (2) tuning agents on domain-specific datasets to improve factual grounding and cross-modal coherence, and (3) incorporating human preference models into evaluation loops to better align automated feedback with human judgment. Although preliminary, \textit{SciTalk} offers foundational insights toward scalable, automatic scientific multimedia generation, highlighting both its promise and current challenges.

\section*{Acknowledgements}
We thank Minnesota NLP group members for providing us with valuable feedback and comments on the initial draft.
We would also like to thank Tripp Dow for his contribution to an earlier prototype of video generation, and Waleed Ammar for initial discussion of the project idea.

\bibliography{colm2025_conference}

\begin{thebibliography}{43}
\providecommand{\natexlab}[1]{#1}
\providecommand{\url}[1]{\texttt{#1}}
\expandafter\ifx\csname urlstyle\endcsname\relax
  \providecommand{\doi}[1]{doi: #1}\else
  \providecommand{\doi}{doi: \begingroup \urlstyle{rm}\Url}\fi

\bibitem[Ahn et~al.(2024)Ahn, Choi, Yu, Kang, and Choi]{ahn-etal-2024-tuning}
Daechul Ahn, Yura Choi, Youngjae Yu, Dongyeop Kang, and Jonghyun Choi.
\newblock Tuning large multimodal models for videos using reinforcement learning from {AI} feedback.
\newblock In Lun-Wei Ku, Andre Martins, and Vivek Srikumar (eds.), \emph{Proceedings of the 62nd Annual Meeting of the Association for Computational Linguistics (Volume 1: Long Papers)}, pp.\  923--940, Bangkok, Thailand, August 2024. Association for Computational Linguistics.
\newblock \doi{10.18653/v1/2024.acl-long.52}.
\newblock URL \url{https://aclanthology.org/2024.acl-long.52/}.

\bibitem[Brooks et~al.(2024)Brooks, Peebles, Holmes, DePue, Guo, Jing, Schnurr, Taylor, Luhman, Luhman, Ng, Wang, and Ramesh]{videoworldsimulators2024}
Tim Brooks, Bill Peebles, Connor Holmes, Will DePue, Yufei Guo, Li~Jing, David Schnurr, Joe Taylor, Troy Luhman, Eric Luhman, Clarence Ng, Ricky Wang, and Aditya Ramesh.
\newblock Video generation models as world simulators.
\newblock 2024.
\newblock URL \url{https://openai.com/research/video-generation-models-as-world-simulators}.

\bibitem[Burns et~al.(2023)Burns, Srinivasan, Ainslie, Brown, Plummer, Saenko, Ni, and Guo]{burns2023wiki}
Andrea Burns, Krishna Srinivasan, Joshua Ainslie, Geoff Brown, Bryan~A. Plummer, Kate Saenko, Jianmo Ni, and Mandy Guo.
\newblock A suite of generative tasks for multi-level multimodal webpage understanding.
\newblock In \emph{The 2023 Conference on Empirical Methods in Natural Language Processing (EMNLP)}, 2023.
\newblock URL \url{https://openreview.net/forum?id=rwcLHjtUmn}.

\bibitem[Chen et~al.(2023)Chen, Sarokin, Lee, Tang, Chang, Kulik, and Grundmann]{Chen_2023_CVPR}
Yu-Hui Chen, Raman Sarokin, Juhyun Lee, Jiuqiang Tang, Chuo-Ling Chang, Andrei Kulik, and Matthias Grundmann.
\newblock Speed is all you need: On-device acceleration of large diffusion models via gpu-aware optimizations.
\newblock In \emph{Proceedings of the IEEE/CVF Conference on Computer Vision and Pattern Recognition (CVPR) Workshops}, pp.\  4651--4655, June 2023.

\bibitem[Choi et~al.(2018)Choi, Oh, and Kweon]{8354295}
Jinsoo Choi, Tae-Hyun Oh, and In~So Kweon.
\newblock Contextually customized video summaries via natural language.
\newblock In \emph{2018 IEEE Winter Conference on Applications of Computer Vision (WACV)}, pp.\  1718--1726, 2018.
\newblock \doi{10.1109/WACV.2018.00191}.

\bibitem[Choi et~al.(2023)Choi, Kang, Lee, and Kim]{10.1145/3544548.3581386}
Yoonseo Choi, Eun~Jeong Kang, Min~Kyung Lee, and Juho Kim.
\newblock Creator-friendly algorithms: Behaviors, challenges, and design opportunities in algorithmic platforms.
\newblock In \emph{Proceedings of the 2023 CHI Conference on Human Factors in Computing Systems}, CHI '23, New York, NY, USA, 2023. Association for Computing Machinery.
\newblock ISBN 9781450394215.
\newblock \doi{10.1145/3544548.3581386}.
\newblock URL \url{https://doi.org/10.1145/3544548.3581386}.

\bibitem[De~Langis et~al.(2024)De~Langis, Koo, and Kang]{de-langis-etal-2024-dynamic}
Karin De~Langis, Ryan Koo, and Dongyeop Kang.
\newblock Dynamic multi-reward weighting for multi-style controllable generation.
\newblock In Yaser Al-Onaizan, Mohit Bansal, and Yun-Nung Chen (eds.), \emph{Proceedings of the 2024 Conference on Empirical Methods in Natural Language Processing}, pp.\  6783--6800, Miami, Florida, USA, November 2024. Association for Computational Linguistics.
\newblock \doi{10.18653/v1/2024.emnlp-main.386}.
\newblock URL \url{https://aclanthology.org/2024.emnlp-main.386/}.

\bibitem[Fu et~al.(2021)Fu, Wang, McDuff, and Song]{Fu2021DOC2PPTAP}
Tsu-Jui Fu, William~Yang Wang, Daniel~J. McDuff, and Yale Song.
\newblock Doc2ppt: Automatic presentation slides generation from scientific documents.
\newblock In \emph{AAAI Conference on Artificial Intelligence}, 2021.
\newblock URL \url{https://api.semanticscholar.org/CorpusID:231719374}.

\bibitem[Ghosh \& Figueroa(2023)Ghosh and Figueroa]{ghosh2023establishing}
Sourojit Ghosh and Andrea Figueroa.
\newblock Establishing tiktok as a platform for informal learning: Evidence from mixed-methods analysis of creators and viewers.
\newblock 2023.
\newblock \doi{10.24251/HICSS.2023.300}.
\newblock URL \url{https://hdl.handle.net/10125/102931}.

\bibitem[He et~al.(2024)He, Song, Huang, Aliaga, and Zhou]{he2024kubrickmultimodalagentcollaborations}
Liu He, Yizhi Song, Hejun Huang, Daniel Aliaga, and Xin Zhou.
\newblock Kubrick: Multimodal agent collaborations for synthetic video generation, 2024.
\newblock URL \url{https://arxiv.org/abs/2408.10453}.

\bibitem[Head et~al.(2021)Head, Lo, Kang, Fok, Skjonsberg, Weld, and Hearst]{head2021augmenting}
Andrew Head, Kyle Lo, Dongyeop Kang, Raymond Fok, Sam Skjonsberg, Daniel~S Weld, and Marti~A Hearst.
\newblock Augmenting scientific papers with just-in-time, position-sensitive definitions of terms and symbols.
\newblock In \emph{Proceedings of the 2021 CHI Conference on Human Factors in Computing Systems}, pp.\  1--18, 2021.

\bibitem[Herman(2023)]{inproceedings}
Laura Herman.
\newblock For who page? tiktok creators’ algorithmic dependencies.
\newblock 10 2023.
\newblock \doi{10.21606/iasdr.2023.576}.

\bibitem[Ho et~al.(2020)Ho, Jain, and Abbeel]{10.5555/3495724.3496298}
Jonathan Ho, Ajay Jain, and Pieter Abbeel.
\newblock Denoising diffusion probabilistic models.
\newblock In \emph{Proceedings of the 34th International Conference on Neural Information Processing Systems}, NIPS '20, Red Hook, NY, USA, 2020. Curran Associates Inc.
\newblock ISBN 9781713829546.

\bibitem[Ho et~al.(2022)Ho, Salimans, Gritsenko, Chan, Norouzi, and Fleet]{NEURIPS2022_39235c56}
Jonathan Ho, Tim Salimans, Alexey Gritsenko, William Chan, Mohammad Norouzi, and David~J Fleet.
\newblock Video diffusion models.
\newblock In S.~Koyejo, S.~Mohamed, A.~Agarwal, D.~Belgrave, K.~Cho, and A.~Oh (eds.), \emph{Advances in Neural Information Processing Systems}, volume~35, pp.\  8633--8646. Curran Associates, Inc., 2022.
\newblock URL \url{https://proceedings.neurips.cc/paper_files/paper/2022/file/39235c56aef13fb05a6adc95eb9d8d66-Paper-Conference.pdf}.

\bibitem[Hu et~al.(2024)Hu, Jiang, Chen, Han, Liao, Chang, and Liang]{hu2024storyagentcustomizedstorytellingvideo}
Panwen Hu, Jin Jiang, Jianqi Chen, Mingfei Han, Shengcai Liao, Xiaojun Chang, and Xiaodan Liang.
\newblock Storyagent: Customized storytelling video generation via multi-agent collaboration, 2024.
\newblock URL \url{https://arxiv.org/abs/2411.04925}.

\bibitem[Huang et~al.(2022)Huang, Shih, and Yang]{10.1145/3477314.3507141}
Hsin-I Huang, Chi-Sheng Shih, and Zi-Lin Yang.
\newblock Automated video editing based on learned styles using lstm-gan.
\newblock In \emph{Proceedings of the 37th ACM/SIGAPP Symposium on Applied Computing}, SAC '22, pp.\  73–80, New York, NY, USA, 2022. Association for Computing Machinery.
\newblock ISBN 9781450387132.
\newblock \doi{10.1145/3477314.3507141}.
\newblock URL \url{https://doi.org/10.1145/3477314.3507141}.

\bibitem[Huang et~al.(2024)Huang, Huang, Ning, Lin, Wang, and Liu]{huang2024genmaccompositionaltexttovideogeneration}
Kaiyi Huang, Yukun Huang, Xuefei Ning, Zinan Lin, Yu~Wang, and Xihui Liu.
\newblock Genmac: Compositional text-to-video generation with multi-agent collaboration.
\newblock 2024.
\newblock URL \url{https://arxiv.org/abs/2412.04440}.

\bibitem[Kang et~al.(2018)Kang, Ammar, Dalvi, van Zuylen, Kohlmeier, Hovy, and Schwartz]{kang-etal-2018-dataset}
Dongyeop Kang, Waleed Ammar, Bhavana Dalvi, Madeleine van Zuylen, Sebastian Kohlmeier, Eduard Hovy, and Roy Schwartz.
\newblock A dataset of peer reviews ({P}eer{R}ead): Collection, insights and {NLP} applications.
\newblock In Marilyn Walker, Heng Ji, and Amanda Stent (eds.), \emph{Proceedings of the 2018 Conference of the North {A}merican Chapter of the Association for Computational Linguistics: Human Language Technologies, Volume 1 (Long Papers)}, pp.\  1647--1661, New Orleans, Louisiana, June 2018. Association for Computational Linguistics.
\newblock \doi{10.18653/v1/N18-1149}.
\newblock URL \url{https://aclanthology.org/N18-1149/}.

\bibitem[Kim et~al.(2024)Kim, Lee, Zhu, Raheja, and Kang]{kim-etal-2024-threads}
Zae~Myung Kim, Kwang Lee, Preston Zhu, Vipul Raheja, and Dongyeop Kang.
\newblock Threads of subtlety: Detecting machine-generated texts through discourse motifs.
\newblock In Lun-Wei Ku, Andre Martins, and Vivek Srikumar (eds.), \emph{Proceedings of the 62nd Annual Meeting of the Association for Computational Linguistics (Volume 1: Long Papers)}, pp.\  5449--5474, Bangkok, Thailand, August 2024. Association for Computational Linguistics.
\newblock \doi{10.18653/v1/2024.acl-long.298}.
\newblock URL \url{https://aclanthology.org/2024.acl-long.298/}.

\bibitem[Klug(2020)]{Klug_2020}
Daniel Klug.
\newblock “it took me almost 30 minutes to practice this.” performance and production practices in dance challenge videos on tiktok.
\newblock August 2020.
\newblock \doi{10.33767/osf.io/j8u9v}.
\newblock URL \url{http://dx.doi.org/10.33767/osf.io/j8u9v}.

\bibitem[Kumar et~al.(2024)Kumar, Kohli, Ghosal, and Ekbal]{kumar-etal-2024-longform}
Sandeep Kumar, Guneet~Singh Kohli, Tirthankar Ghosal, and Asif Ekbal.
\newblock Longform multimodal lay summarization of scientific papers: Towards automatically generating science blogs from research articles.
\newblock In Nicoletta Calzolari, Min-Yen Kan, Veronique Hoste, Alessandro Lenci, Sakriani Sakti, and Nianwen Xue (eds.), \emph{Proceedings of the 2024 Joint International Conference on Computational Linguistics, Language Resources and Evaluation (LREC-COLING 2024)}, pp.\  10790--10801, Torino, Italia, May 2024. ELRA and ICCL.
\newblock URL \url{https://aclanthology.org/2024.lrec-main.942/}.

\bibitem[Li et~al.(2024)Li, Wang, Xu, Wang, Feng, Kong, and Liu]{li-etal-2024-multimodal-arxiv}
Lei Li, Yuqi Wang, Runxin Xu, Peiyi Wang, Xiachong Feng, Lingpeng Kong, and Qi~Liu.
\newblock Multimodal {A}r{X}iv: A dataset for improving scientific comprehension of large vision-language models.
\newblock In Lun-Wei Ku, Andre Martins, and Vivek Srikumar (eds.), \emph{Proceedings of the 62nd Annual Meeting of the Association for Computational Linguistics (Volume 1: Long Papers)}, pp.\  14369--14387, Bangkok, Thailand, August 2024. Association for Computational Linguistics.
\newblock \doi{10.18653/v1/2024.acl-long.775}.
\newblock URL \url{https://aclanthology.org/2024.acl-long.775}.

\bibitem[Lu et~al.(2024)Lu, CAI, Li, Qin, and Li]{lu2024improve}
Kexin Lu, Yuxi CAI, Lan Li, Dafei Qin, and Guodong Li.
\newblock Improve temporal consistency in diffusion models through noise correlations, 2024.
\newblock URL \url{https://openreview.net/forum?id=59nCKifDtm}.

\bibitem[Morris \& Rush(2025)Morris and Rush]{morris2025contextual}
John~Xavier Morris and Alexander~M Rush.
\newblock Contextual document embeddings.
\newblock In \emph{The Thirteenth International Conference on Learning Representations}, 2025.
\newblock URL \url{https://openreview.net/forum?id=Wqsk3FbD6D}.

\bibitem[OpenAI et~al.(2024)OpenAI, :, Hurst, Lerer, Goucher, Perelman, Ramesh, Clark, Ostrow, Welihinda, Hayes, Radford, Madry, Baker-Whitcomb, Beutel, Borzunov, Carney, Chow, Kirillov, Nichol, Paino, Renzin, Passos, Kirillov, Christakis, Conneau, Kamali, Jabri, Moyer, Tam, Crookes, Tootoochian, Tootoonchian, Kumar, Vallone, Karpathy, Braunstein, Cann, Codispoti, Galu, Kondrich, Tulloch, Mishchenko, Baek, Jiang, Pelisse, Woodford, Gosalia, Dhar, Pantuliano, Nayak, Oliver, Zoph, Ghorbani, Leimberger, Rossen, Sokolowsky, Wang, Zweig, Hoover, Samic, McGrew, Spero, Giertler, Cheng, Lightcap, Walkin, Quinn, Guarraci, Hsu, Kellogg, Eastman, Lugaresi, Wainwright, Bassin, Hudson, Chu, Nelson, Li, Shern, Conger, Barette, Voss, Ding, Lu, Zhang, Beaumont, Hallacy, Koch, Gibson, Kim, Choi, McLeavey, Hesse, Fischer, Winter, Czarnecki, Jarvis, Wei, Koumouzelis, Sherburn, Kappler, Levin, Levy, Carr, Farhi, Mely, Robinson, Sasaki, Jin, Valladares, Tsipras, Li, Nguyen, Findlay, Oiwoh, Wong, Asdar, Proehl, Yang, Antonow,
  Kramer, Peterson, Sigler, Wallace, Brevdo, Mays, Khorasani, Such, Raso, Zhang, von Lohmann, Sulit, Goh, Oden, Salmon, Starace, Brockman, Salman, Bao, Hu, Wong, Wang, Schmidt, Whitney, Jun, Kirchner, de~Oliveira~Pinto, Ren, Chang, Chung, Kivlichan, O'Connell, O'Connell, Osband, Silber, Sohl, Okuyucu, Lan, Kostrikov, Sutskever, Kanitscheider, Gulrajani, Coxon, Menick, Pachocki, Aung, Betker, Crooks, Lennon, Kiros, Leike, Park, Kwon, Phang, Teplitz, Wei, Wolfe, Chen, Harris, Varavva, Lee, Shieh, Lin, Yu, Weng, Tang, Yu, Jang, Candela, Beutler, Landers, Parish, Heidecke, Schulman, Lachman, McKay, Uesato, Ward, Kim, Huizinga, Sitkin, Kraaijeveld, Gross, Kaplan, Snyder, Achiam, Jiao, Lee, Zhuang, Harriman, Fricke, Hayashi, Singhal, Shi, Karthik, Wood, Rimbach, Hsu, Nguyen, Gu-Lemberg, Button, Liu, Howe, Muthukumar, Luther, Ahmad, Kai, Itow, Workman, Pathak, Chen, Jing, Guy, Fedus, Zhou, Mamitsuka, Weng, McCallum, Held, Ouyang, Feuvrier, Zhang, Kondraciuk, Kaiser, Hewitt, Metz, Doshi, Aflak, Simens, Boyd,
  Thompson, Dukhan, Chen, Gray, Hudnall, Zhang, Aljubeh, Litwin, Zeng, Johnson, Shetty, Gupta, Shah, Yatbaz, Yang, Zhong, Glaese, Chen, Janner, Lampe, Petrov, Wu, Wang, Fradin, Pokrass, Castro, de~Castro, Pavlov, Brundage, Wang, Khan, Murati, Bavarian, Lin, Yesildal, Soto, Gimelshein, Cone, Staudacher, Summers, LaFontaine, Chowdhury, Ryder, Stathas, Turley, Tezak, Felix, Kudige, Keskar, Deutsch, Bundick, Puckett, Nachum, Okelola, Boiko, Murk, Jaffe, Watkins, Godement, Campbell-Moore, Chao, McMillan, Belov, Su, Bak, Bakkum, Deng, Dolan, Hoeschele, Welinder, Tillet, Pronin, Tillet, Dhariwal, Yuan, Dias, Lim, Arora, Troll, Lin, Lopes, Puri, Miyara, Leike, Gaubert, Zamani, Wang, Donnelly, Honsby, Smith, Sahai, Ramchandani, Huet, Carmichael, Zellers, Chen, Chen, Nigmatullin, Cheu, Jain, Altman, Schoenholz, Toizer, Miserendino, Agarwal, Culver, Ethersmith, Gray, Grove, Metzger, Hermani, Jain, Zhao, Wu, Jomoto, Wu, Shuaiqi, Xia, Phene, Papay, Narayanan, Coffey, Lee, Hall, Balaji, Broda, Stramer, Xu, Gogineni,
  Christianson, Sanders, Patwardhan, Cunninghman, Degry, Dimson, Raoux, Shadwell, Zheng, Underwood, Markov, Sherbakov, Rubin, Stasi, Kaftan, Heywood, Peterson, Walters, Eloundou, Qi, Moeller, Monaco, Kuo, Fomenko, Chang, Zheng, Zhou, Manassra, Sheu, Zaremba, Patil, Qian, Kim, Cheng, Zhang, He, Zhang, Jin, Dai, and Malkov]{openai2024gpt4ocard}
OpenAI, :, Aaron Hurst, Adam Lerer, Adam~P. Goucher, Adam Perelman, Aditya Ramesh, Aidan Clark, AJ~Ostrow, Akila Welihinda, Alan Hayes, Alec Radford, Aleksander Madry, Alex Baker-Whitcomb, Alex Beutel, Alex Borzunov, Alex Carney, Alex Chow, Alex Kirillov, Alex Nichol, Alex Paino, Alex Renzin, Alex~Tachard Passos, Alexander Kirillov, Alexi Christakis, Alexis Conneau, Ali Kamali, Allan Jabri, Allison Moyer, Allison Tam, Amadou Crookes, Amin Tootoochian, Amin Tootoonchian, Ananya Kumar, Andrea Vallone, Andrej Karpathy, Andrew Braunstein, Andrew Cann, Andrew Codispoti, Andrew Galu, Andrew Kondrich, Andrew Tulloch, Andrey Mishchenko, Angela Baek, Angela Jiang, Antoine Pelisse, Antonia Woodford, Anuj Gosalia, Arka Dhar, Ashley Pantuliano, Avi Nayak, Avital Oliver, Barret Zoph, Behrooz Ghorbani, Ben Leimberger, Ben Rossen, Ben Sokolowsky, Ben Wang, Benjamin Zweig, Beth Hoover, Blake Samic, Bob McGrew, Bobby Spero, Bogo Giertler, Bowen Cheng, Brad Lightcap, Brandon Walkin, Brendan Quinn, Brian Guarraci, Brian Hsu,
  Bright Kellogg, Brydon Eastman, Camillo Lugaresi, Carroll Wainwright, Cary Bassin, Cary Hudson, Casey Chu, Chad Nelson, Chak Li, Chan~Jun Shern, Channing Conger, Charlotte Barette, Chelsea Voss, Chen Ding, Cheng Lu, Chong Zhang, Chris Beaumont, Chris Hallacy, Chris Koch, Christian Gibson, Christina Kim, Christine Choi, Christine McLeavey, Christopher Hesse, Claudia Fischer, Clemens Winter, Coley Czarnecki, Colin Jarvis, Colin Wei, Constantin Koumouzelis, Dane Sherburn, Daniel Kappler, Daniel Levin, Daniel Levy, David Carr, David Farhi, David Mely, David Robinson, David Sasaki, Denny Jin, Dev Valladares, Dimitris Tsipras, Doug Li, Duc~Phong Nguyen, Duncan Findlay, Edede Oiwoh, Edmund Wong, Ehsan Asdar, Elizabeth Proehl, Elizabeth Yang, Eric Antonow, Eric Kramer, Eric Peterson, Eric Sigler, Eric Wallace, Eugene Brevdo, Evan Mays, Farzad Khorasani, Felipe~Petroski Such, Filippo Raso, Francis Zhang, Fred von Lohmann, Freddie Sulit, Gabriel Goh, Gene Oden, Geoff Salmon, Giulio Starace, Greg Brockman, Hadi
  Salman, Haiming Bao, Haitang Hu, Hannah Wong, Haoyu Wang, Heather Schmidt, Heather Whitney, Heewoo Jun, Hendrik Kirchner, Henrique~Ponde de~Oliveira~Pinto, Hongyu Ren, Huiwen Chang, Hyung~Won Chung, Ian Kivlichan, Ian O'Connell, Ian O'Connell, Ian Osband, Ian Silber, Ian Sohl, Ibrahim Okuyucu, Ikai Lan, Ilya Kostrikov, Ilya Sutskever, Ingmar Kanitscheider, Ishaan Gulrajani, Jacob Coxon, Jacob Menick, Jakub Pachocki, James Aung, James Betker, James Crooks, James Lennon, Jamie Kiros, Jan Leike, Jane Park, Jason Kwon, Jason Phang, Jason Teplitz, Jason Wei, Jason Wolfe, Jay Chen, Jeff Harris, Jenia Varavva, Jessica~Gan Lee, Jessica Shieh, Ji~Lin, Jiahui Yu, Jiayi Weng, Jie Tang, Jieqi Yu, Joanne Jang, Joaquin~Quinonero Candela, Joe Beutler, Joe Landers, Joel Parish, Johannes Heidecke, John Schulman, Jonathan Lachman, Jonathan McKay, Jonathan Uesato, Jonathan Ward, Jong~Wook Kim, Joost Huizinga, Jordan Sitkin, Jos Kraaijeveld, Josh Gross, Josh Kaplan, Josh Snyder, Joshua Achiam, Joy Jiao, Joyce Lee, Juntang
  Zhuang, Justyn Harriman, Kai Fricke, Kai Hayashi, Karan Singhal, Katy Shi, Kavin Karthik, Kayla Wood, Kendra Rimbach, Kenny Hsu, Kenny Nguyen, Keren Gu-Lemberg, Kevin Button, Kevin Liu, Kiel Howe, Krithika Muthukumar, Kyle Luther, Lama Ahmad, Larry Kai, Lauren Itow, Lauren Workman, Leher Pathak, Leo Chen, Li~Jing, Lia Guy, Liam Fedus, Liang Zhou, Lien Mamitsuka, Lilian Weng, Lindsay McCallum, Lindsey Held, Long Ouyang, Louis Feuvrier, Lu~Zhang, Lukas Kondraciuk, Lukasz Kaiser, Luke Hewitt, Luke Metz, Lyric Doshi, Mada Aflak, Maddie Simens, Madelaine Boyd, Madeleine Thompson, Marat Dukhan, Mark Chen, Mark Gray, Mark Hudnall, Marvin Zhang, Marwan Aljubeh, Mateusz Litwin, Matthew Zeng, Max Johnson, Maya Shetty, Mayank Gupta, Meghan Shah, Mehmet Yatbaz, Meng~Jia Yang, Mengchao Zhong, Mia Glaese, Mianna Chen, Michael Janner, Michael Lampe, Michael Petrov, Michael Wu, Michele Wang, Michelle Fradin, Michelle Pokrass, Miguel Castro, Miguel Oom~Temudo de~Castro, Mikhail Pavlov, Miles Brundage, Miles Wang, Minal
  Khan, Mira Murati, Mo~Bavarian, Molly Lin, Murat Yesildal, Nacho Soto, Natalia Gimelshein, Natalie Cone, Natalie Staudacher, Natalie Summers, Natan LaFontaine, Neil Chowdhury, Nick Ryder, Nick Stathas, Nick Turley, Nik Tezak, Niko Felix, Nithanth Kudige, Nitish Keskar, Noah Deutsch, Noel Bundick, Nora Puckett, Ofir Nachum, Ola Okelola, Oleg Boiko, Oleg Murk, Oliver Jaffe, Olivia Watkins, Olivier Godement, Owen Campbell-Moore, Patrick Chao, Paul McMillan, Pavel Belov, Peng Su, Peter Bak, Peter Bakkum, Peter Deng, Peter Dolan, Peter Hoeschele, Peter Welinder, Phil Tillet, Philip Pronin, Philippe Tillet, Prafulla Dhariwal, Qiming Yuan, Rachel Dias, Rachel Lim, Rahul Arora, Rajan Troll, Randall Lin, Rapha~Gontijo Lopes, Raul Puri, Reah Miyara, Reimar Leike, Renaud Gaubert, Reza Zamani, Ricky Wang, Rob Donnelly, Rob Honsby, Rocky Smith, Rohan Sahai, Rohit Ramchandani, Romain Huet, Rory Carmichael, Rowan Zellers, Roy Chen, Ruby Chen, Ruslan Nigmatullin, Ryan Cheu, Saachi Jain, Sam Altman, Sam Schoenholz, Sam
  Toizer, Samuel Miserendino, Sandhini Agarwal, Sara Culver, Scott Ethersmith, Scott Gray, Sean Grove, Sean Metzger, Shamez Hermani, Shantanu Jain, Shengjia Zhao, Sherwin Wu, Shino Jomoto, Shirong Wu, Shuaiqi, Xia, Sonia Phene, Spencer Papay, Srinivas Narayanan, Steve Coffey, Steve Lee, Stewart Hall, Suchir Balaji, Tal Broda, Tal Stramer, Tao Xu, Tarun Gogineni, Taya Christianson, Ted Sanders, Tejal Patwardhan, Thomas Cunninghman, Thomas Degry, Thomas Dimson, Thomas Raoux, Thomas Shadwell, Tianhao Zheng, Todd Underwood, Todor Markov, Toki Sherbakov, Tom Rubin, Tom Stasi, Tomer Kaftan, Tristan Heywood, Troy Peterson, Tyce Walters, Tyna Eloundou, Valerie Qi, Veit Moeller, Vinnie Monaco, Vishal Kuo, Vlad Fomenko, Wayne Chang, Weiyi Zheng, Wenda Zhou, Wesam Manassra, Will Sheu, Wojciech Zaremba, Yash Patil, Yilei Qian, Yongjik Kim, Youlong Cheng, Yu~Zhang, Yuchen He, Yuchen Zhang, Yujia Jin, Yunxing Dai, and Yury Malkov.
\newblock Gpt-4o system card, 2024.
\newblock URL \url{https://arxiv.org/abs/2410.21276}.

\bibitem[Ren et~al.(2024)Ren, Yang, Zhang, Wei, Du, Huang, and Chen]{ren2024consistiv}
Weiming Ren, Huan Yang, Ge~Zhang, Cong Wei, Xinrun Du, Wenhao Huang, and Wenhu Chen.
\newblock Consisti2v: Enhancing visual consistency for image-to-video generation.
\newblock \emph{Transactions on Machine Learning Research}, 2024.
\newblock ISSN 2835-8856.
\newblock URL \url{https://openreview.net/forum?id=vqniLmUDvj}.

\bibitem[{Research!America}(2020)]{researchamerica2020flashtalks}
{Research!America}.
\newblock Flash talks competition, September 2020.
\newblock URL \url{https://www.youtube.com/watch?v=2MLEq6-zk1A}.
\newblock Accessed: March 27, 2025.

\bibitem[Shorten et~al.(2025)Shorten, Pierse, Smith, D'Oosterlinck, Celik, Cardenas, Monigatti, Hasan, Schmuhl, Williams, Kesiraju, and van Luijt]{shorten2025queryingdatabasesfunctioncalling}
Connor Shorten, Charles Pierse, Thomas~Benjamin Smith, Karel D'Oosterlinck, Tuana Celik, Erika Cardenas, Leonie Monigatti, Mohd~Shukri Hasan, Edward Schmuhl, Daniel Williams, Aravind Kesiraju, and Bob van Luijt.
\newblock Querying databases with function calling, 2025.
\newblock URL \url{https://arxiv.org/abs/2502.00032}.

\bibitem[Singer et~al.(2023)Singer, Polyak, Hayes, Yin, An, Zhang, Hu, Yang, Ashual, Gafni, Parikh, Gupta, and Taigman]{singer2023makeavideo}
Uriel Singer, Adam Polyak, Thomas Hayes, Xi~Yin, Jie An, Songyang Zhang, Qiyuan Hu, Harry Yang, Oron Ashual, Oran Gafni, Devi Parikh, Sonal Gupta, and Yaniv Taigman.
\newblock Make-a-video: Text-to-video generation without text-video data.
\newblock In \emph{The Eleventh International Conference on Learning Representations}, 2023.
\newblock URL \url{https://openreview.net/forum?id=nJfylDvgzlq}.

\bibitem[Soni et~al.(2025)Soni, Venkataraman, Chandra, Fischmeister, Liang, Dai, and Yang]{soni2025videoagent}
Achint Soni, Sreyas Venkataraman, Abhranil Chandra, Sebastian Fischmeister, Percy Liang, Bo~Dai, and Sherry Yang.
\newblock Videoagent: Self-improving video generation, 2025.
\newblock URL \url{https://openreview.net/forum?id=JaRihIHbZm}.

\bibitem[Svennevig et~al.(2024)Svennevig, Hicks, Forbriger, Lecocq, Widmer-Schnidrig, Mangeney, Hibert, Korsgaard, Lucas, Satriano, Anthony, Mordret, Schippkus, Rysgaard, Boone, Gibbons, Cook, Glimsdal, Løvholt, Noten, Assink, Marboeuf, Lomax, Vanneste, Taira, Spagnolo, Plaen, Koelemeijer, Ebeling, Cannata, Harcourt, Cornwell, Caudron, Poli, Bernard, Larose, Stutzmann, Voss, Lund, Cannavo, Castro-Díaz, Chaves, Dahl-Jensen, Dias, Déprez, Develter, Dreger, Evers, Fernández-Nieto, Ferreira, Funning, Gabriel, Hendrickx, Kafka, Keiding, Kerby, Khan, Dideriksen, Lamb, Larsen, Lipovsky, Magdalena, Malet, Myrup, Rivera, Ruiz-Castillo, Wetter, and Wirtz]{doi:10.1126/science.adm9247}
Kristian Svennevig, Stephen~P. Hicks, Thomas Forbriger, Thomas Lecocq, Rudolf Widmer-Schnidrig, Anne Mangeney, Clément Hibert, Niels~J. Korsgaard, Antoine Lucas, Claudio Satriano, Robert~E. Anthony, Aurélien Mordret, Sven Schippkus, Søren Rysgaard, Wieter Boone, Steven~J. Gibbons, Kristen~L. Cook, Sylfest Glimsdal, Finn Løvholt, Koen~Van Noten, Jelle~D. Assink, Alexis Marboeuf, Anthony Lomax, Kris Vanneste, Taka’aki Taira, Matteo Spagnolo, Raphael~De Plaen, Paula Koelemeijer, Carl Ebeling, Andrea Cannata, William~D. Harcourt, David~G. Cornwell, Corentin Caudron, Piero Poli, Pascal Bernard, Eric Larose, Eleonore Stutzmann, Peter~H. Voss, Bjorn Lund, Flavio Cannavo, Manuel~J. Castro-Díaz, Esteban Chaves, Trine Dahl-Jensen, Nicolas De~Pinho Dias, Aline Déprez, Roeland Develter, Douglas Dreger, Läslo~G. Evers, Enrique~D. Fernández-Nieto, Ana M.~G. Ferreira, Gareth Funning, Alice-Agnes Gabriel, Marc Hendrickx, Alan~L. Kafka, Marie Keiding, Jeffrey Kerby, Shfaqat~A. Khan, Andreas~Kjær Dideriksen,
  Oliver~D. Lamb, Tine~B. Larsen, Bradley Lipovsky, Ikha Magdalena, Jean-Philippe Malet, Mikkel Myrup, Luis Rivera, Eugenio Ruiz-Castillo, Selina Wetter, and Bastien Wirtz.
\newblock A rockslide-generated tsunami in a greenland fjord rang earth for 9 days.
\newblock \emph{Science}, 385\penalty0 (6714):\penalty0 1196--1205, 2024.
\newblock \doi{10.1126/science.adm9247}.
\newblock URL \url{https://www.science.org/doi/abs/10.1126/science.adm9247}.

\bibitem[Tu et~al.(2024)Tu, Sun, Jin, Liao, Huang, and Tao]{tu2024spagentadaptivetaskdecomposition}
Rong-Cheng Tu, Wenhao Sun, Zhao Jin, Jingyi Liao, Jiaxing Huang, and Dacheng Tao.
\newblock Spagent: Adaptive task decomposition and model selection for general video generation and editing, 2024.
\newblock URL \url{https://arxiv.org/abs/2411.18983}.

\bibitem[Voleti et~al.(2022)Voleti, Jolicoeur-Martineau, and Pal]{voleti2022MCVD}
Vikram Voleti, Alexia Jolicoeur-Martineau, and Christopher Pal.
\newblock Mcvd: Masked conditional video diffusion for prediction, generation, and interpolation.
\newblock In \emph{(NeurIPS) Advances in Neural Information Processing Systems}, 2022.
\newblock URL \url{https://arxiv.org/abs/2205.09853}.

\bibitem[Wang et~al.(2024)Wang, Du, Zhao, Yuan, Wang, Liang, Zhao, Lu, Li, Gao, Tu, and Guo]{wang2024aesopagentagentdrivenevolutionarystorytovideo}
Jiuniu Wang, Zehua Du, Yuyuan Zhao, Bo~Yuan, Kexiang Wang, Jian Liang, Yaxi Zhao, Yihen Lu, Gengliang Li, Junlong Gao, Xin Tu, and Zhenyu Guo.
\newblock Aesopagent: Agent-driven evolutionary system on story-to-video production, 2024.
\newblock URL \url{https://arxiv.org/abs/2403.07952}.

\bibitem[Wang et~al.(2025)Wang, Lee, Volkov, Chau, and Kang]{wang2025scholawrite}
Linghe Wang, Minhwa Lee, Ross Volkov, Luan~Tuyen Chau, and Dongyeop Kang.
\newblock Scholawrite: A dataset of end-to-end scholarly writing process.
\newblock \emph{arXiv preprint arXiv:2502.02904}, 2025.

\bibitem[W\"{u}nsche et~al.(2023)W\"{u}nsche, Koesten, M\"{o}ller, and Chen]{10.1145/3573381.3596157}
Katharina W\"{u}nsche, Laura Koesten, Torsten M\"{o}ller, and Jian Chen.
\newblock Supporting video authoring for communication of research results.
\newblock In \emph{Proceedings of the 2023 ACM International Conference on Interactive Media Experiences}, IMX '23, pp.\  47–59, New York, NY, USA, 2023. Association for Computing Machinery.
\newblock ISBN 9798400700286.
\newblock \doi{10.1145/3573381.3596157}.
\newblock URL \url{https://doi.org/10.1145/3573381.3596157}.

\bibitem[Xu et~al.(2025)Xu, Wang, Wang, Li, Shi, Yang, Wang, Hu, Yu, and Zhang]{xu2025filmagent}
Zhenran Xu, Longyue Wang, Jifang Wang, Zhouyi Li, Senbao Shi, Xue Yang, Yiyu Wang, Baotian Hu, Jun Yu, and Min Zhang.
\newblock Filmagent: A multi-agent framework for end-to-end film automation in virtual 3d spaces, 2025.
\newblock URL \url{https://arxiv.org/abs/2501.12909}.

\bibitem[Yan et~al.(2025)Yan, Hou, Xiao, Zhang, and Wang]{10670469}
Youfu Yan, Yu~Hou, Yongkang Xiao, Rui Zhang, and Qianwen Wang.
\newblock { KNowNEt:Guided Health Information Seeking from LLMs via Knowledge Graph Integration }.
\newblock \emph{IEEE Transactions on Visualization \& Computer Graphics}, 31\penalty0 (01):\penalty0 547--557, January 2025.
\newblock ISSN 1941-0506.
\newblock \doi{10.1109/TVCG.2024.3456364}.
\newblock URL \url{https://doi.ieeecomputersociety.org/10.1109/TVCG.2024.3456364}.

\bibitem[Yang et~al.(2024)Yang, Yang, Zhang, Hui, Zheng, Yu, Li, Liu, Huang, Wei, Lin, Yang, Tu, Zhang, Yang, Yang, Zhou, Lin, Dang, Lu, Bao, Yang, Yu, Li, Xue, Zhang, Zhu, Men, Lin, Li, Xia, Ren, Ren, Fan, Su, Zhang, Wan, Liu, Cui, Zhang, and Qiu]{qwen2.5}
An~Yang, Baosong Yang, Beichen Zhang, Binyuan Hui, Bo~Zheng, Bowen Yu, Chengyuan Li, Dayiheng Liu, Fei Huang, Haoran Wei, Huan Lin, Jian Yang, Jianhong Tu, Jianwei Zhang, Jianxin Yang, Jiaxi Yang, Jingren Zhou, Junyang Lin, Kai Dang, Keming Lu, Keqin Bao, Kexin Yang, Le~Yu, Mei Li, Mingfeng Xue, Pei Zhang, Qin Zhu, Rui Men, Runji Lin, Tianhao Li, Tingyu Xia, Xingzhang Ren, Xuancheng Ren, Yang Fan, Yang Su, Yichang Zhang, Yu~Wan, Yuqiong Liu, Zeyu Cui, Zhenru Zhang, and Zihan Qiu.
\newblock Qwen2.5 technical report.
\newblock \emph{arXiv preprint arXiv:2412.15115}, 2024.

\bibitem[Yu et~al.(2022)Yu, Tack, Mo, Kim, Kim, Ha, and Shin]{digan}
Sihyun Yu, Jihoon Tack, Sangwoo Mo, Hyunsu Kim, Junho Kim, Jung-Woo Ha, and Jinwoo Shin.
\newblock Generating videos with dynamics-aware implicit generative adversarial networks.
\newblock In \emph{International Conference on Learning Representations}, 2022.
\newblock URL \url{https://openreview.net/forum?id=Czsdv-S4-w9}.

\bibitem[Yuan et~al.(2024)Yuan, Liu, Cao, Sun, Jia, Chen, Li, Lin, Yuan, He, Wang, Ye, and Sun]{yuan2024moraenablinggeneralistvideo}
Zhengqing Yuan, Yixin Liu, Yihan Cao, Weixiang Sun, Haolong Jia, Ruoxi Chen, Zhaoxu Li, Bin Lin, Li~Yuan, Lifang He, Chi Wang, Yanfang Ye, and Lichao Sun.
\newblock Mora: Enabling generalist video generation via a multi-agent framework, 2024.
\newblock URL \url{https://arxiv.org/abs/2403.13248}.

\bibitem[Zhang et~al.(2024)Zhang, Li, Liu, Lee, Gui, Fu, Feng, Liu, and Li]{zhang2024llavanextvideo}
Yuanhan Zhang, Bo~Li, haotian Liu, Yong~jae Lee, Liangke Gui, Di~Fu, Jiashi Feng, Ziwei Liu, and Chunyuan Li.
\newblock Llava-next: A strong zero-shot video understanding model, April 2024.
\newblock URL \url{https://llava-vl.github.io/blog/2024-04-30-llava-next-video/}.

\bibitem[Zulko(2014)]{zulko2014}
Zulko.
\newblock Moviepy: a python library for video editing.
\newblock \url{https://github.com/charlespwd/project-title}, 2014.

\end{thebibliography}
\bibliographystyle{colm2025_conference}

\appendix
\section{Appendix}

\begin{tiny}
\begin{longtable}{p{1.1cm}|p{1.1cm}|p{1.1cm}|p{1.1cm}|p{7cm}}
\toprule
\textbf{Stage} & \textbf{Agent} & \textbf{Role} & \textbf{Output} & \textbf{Example Initial Prompt} \\ \midrule
\endhead

Pre \newline processing & - & Extract \newline material & Images, Paper Text, Captions & - \\
\midrule

Planning & Flashtalk \newline Generator & Generate video \newline outline & Flashtalk JSON file & You are an engaging, curiosity-driven assistant specializing in crafting captivating 1-minute flash talks based on academic papers. Instead of merely summarizing, your goal is to spark genuine curiosity by opening with an aggressive, intriguing hook or provocative question directly related to the paper's core message. Avoid excessive detail; your objective is to excite curiosity and motivate the audience to seek out the full paper themselves. Actively incorporate relevant images and figures selectively and strategically to amplify curiosity, assigning at least one image to each section. Follow this structure: \newline 1. Aggressive Hook: Begin immediately with a provocative question or surprising statement related to the paper's most striking or controversial finding. \newline 2. Brief Context: Provide minimal context to frame why this question or finding is impactful or significant. \newline 3. Intriguing Teaser: Hint at the method or result without giving everything away. Mention or show one compelling figure if available. \newline 4. Call to Action: Conclude by strongly encouraging viewers to explore the full paper themselves, highlighting that more surprising insights await. Maintain a dynamic, energetic, and conversational tone suitable for quickly capturing attention. \\
\cmidrule{2-5}

 & Sceneplan \newline Generator & Define detailed \newline sub-scenes & Sceneplan JSON file & You are an expert in designing engaging and visually appealing short video scripts for academic content. Your task is to generate a detailed sequence of scenes for a 1-minute video based on a flash talk summary of a research paper. The flash talk is divided into several sections: \{ sections \} Each section should be split into several sub-scenes, containing a description, image sources, a duration, and an avatar. A sub-scene should have the same visual sources such as images and an avatar. In other words, a sub-scene is delivering the semantically the same message. Audio files for each section are already generated, with the following durations: \{ section\_durations \} Each section must contain 2 to 4 sub-scenes, with each sub-scene lasting between 2 to 8 seconds, ensuring the total duration should fit the section's allocated time. Provide a description for each sub-scene, which will serve as the basis for video creation. Ensure the description should be implementable within the tool descriptions. Ensure a smooth flow between scenes with seamless transitions and avoid abrupt cuts. Focus on a clean, spacious, and visually engaging design with minimal on-screen text. Use color schemes and effects that enhance the content without causing distractions. The final video should be accessible and captivating, even for viewers unfamiliar with the subject matter. \\
\midrule

Editing & Background \newline Assistant & Select suitable \newline backgrounds & Selected Background Image & - \\
\cmidrule{2-5}

 & Text \newline Assistant & Generate text \newline on a scene & Text Components JSON file & You are Text Assistant, a skilled expert in extracting and organizing text components from descriptions to enhance visual storytelling. Your task is to identify relevant text sources, define their start and end points, and ensure their style align with the tone and purpose of the scene. Approach this task with the expertise of a graphic designer, structuring text elements cohesively to support the scene's narrative. \\
\cmidrule{2-5}


 & Layout \newline Allocator & Allocate \newline visual components  & Visual Components Layout JSON File & You are Layout Allocator, an expert in arranging visual components (text, image, figure, avatar, etc.) for a video sub-scene. Your primary responsibility is to determine precise positions (top and left) for **all** visual elements in the layout, ensuring alignment with the sub-scene description. You are provided with an HTML structure containing div elements, each representing a visual component. Your output must include **all** div components' positions, ensuring no component is omitted. \\
 \midrule

Feedback & Flashtalk \newline Feedback \newline Agent & Provide \newline feedback on \newline video outline & Feedback Summary Text File & You are Flashtalk Feedback Agent, an advanced multi-modal evaluator specializing in assessing 1-minute curiosity-driven flash talks, which serve as engaging overviews to drive audience interest in video generation. Rate briefly using these metrics:

1. Curiosity (1-5): Does the hook immediately capture audience interest?
\newline 2. Clarity (1-5): Is the message clear and easy to follow?
\newline 3. Effectiveness (1-5): Does the flash talk motivate viewers to explore the full content?

Each metric is rated on a scale of 1 to 5, where:
  \newline - 1 = Poor (Significant issues, unclear or misaligned)
  \newline - 3 = Adequate (Somewhat effective but needs improvement)
  \newline - 5 = Excellent (Clear, engaging, and well-aligned)

Your feedback should include:
  \newline - Strengths of the flashtalk and video.
  \newline - Areas for improvement in curiosity, clarity, and effectiveness.
  \newline - Actionable suggestions. \\
\cmidrule{2-5}

 & Sceneplan \newline Feedback \newline Agent & Provide \newline feedback on \newline sceneplan & Sceneplan Feedback Text File & You are a Sceneplan Feedback Agent, an expert in evaluating the structure, coherence, and effectiveness of scene sequences in short academic videos. Your role is to assess the Sceneplan generated for a 1-minute video based on a flash talk and provide detailed feedback on its alignment with the intended content and overall storytelling impact.

The Sceneplan consists of multiple sections, each containing sub-scenes with descriptions, images, and duration constraints. These sub-scenes should be visually engaging, well-paced, and smoothly transitioned, ensuring a logical flow and effective communication.

Evaluate the Sceneplan using the following three key metrics:

1. Narrative Coherence:
  \newline - Does the sequence of sub-scenes follow a logical flow?
  \newline - Are transitions smooth, avoiding abrupt shifts between scenes?
  \newline - Does the structure maintain audience engagement throughout the video?

2. Visual Relevance \& Clarity:
  \newline - Do the selected images and visuals align with the sub-scene descriptions?
  \newline - Are there any unclear or ambiguous visual choices?
  \newline - Does each sub-scene effectively illustrate the corresponding narration?

3. Timing \& Pacing:
  \newline - Are the durations of sub-scenes appropriately distributed?
  \newline - Do any sections feel rushed or overly extended?
  \newline - Is the pacing consistent, maintaining viewer engagement?

Each metric is rated on a scale of **1 to 5**, where:
  \newline - 1 = Poor (Significant issues, unclear or misaligned)
  \newline - 3 = Adequate (Somewhat effective but needs improvement)
  \newline - 5 = Excellent (Clear, engaging, and well-aligned)

Your feedback should include:
  \newline - Strengths of the Sceneplan in terms of structure, visuals, and pacing.
  \newline - Areas for improvement, focusing on logical flow, clarity, and engagement.
  \newline - Actionable suggestions to enhance scene transitions, visual storytelling, and timing adjustments. \\
\cmidrule{2-5}

 & Text \newline Feedback \newline Agent & Provide \newline feedback on \newline text \newline components & Text Feedback Summary Text File & You are Text Feedback Agent, an expert evaluator for on-screen text components extracted by Text Assistant to enhance visual storytelling.

Text Assistant extracts key phrases from scene descriptions to highlight main ideas clearly and cohesively, structuring them as effective on-screen text for visual storytelling.

Evaluate concisely using these metrics:
1. Clarity (1-5):
  \newline - Are the text components clear, concise, and logically structured?
  \newline - Do they enhance scene understanding without clutter?
\newline 2. Key Information Coverage (1-5):
  \newline - Do the extracted texts effectively summarize core ideas from the scene?
\newline 3. Timing and Alignment (1-5):
  \newline - Are text components timed appropriately with the audio narration and visuals?
  \newline - Do the texts appear and disappear naturally to support visual storytelling?

Provide brief feedback clearly highlighting:
  \newline - Main strengths
  \newline - Areas for improvement
  \newline - Short, actionable suggestions. \\
\cmidrule{2-5}


& Reflection \newline Agent & Integrate \newline feedback \newline into \newline prompts & New Prompt JSON File & You are an expert in refining and restructuring flash talk prompts based on feedback from a multi-modal large language model (MLLM). \newline Your goal is to integrate suggested improvements while optimizing the clarity, engagement, and accuracy of the generated flash talk. \newline You have autonomy to dynamically determine the most suitable sections for structuring the flash talk rather than adhering to a fixed format. \newline \newline Your approach: \newline 1. Analyze the MLLM feedback and identify key actionable points that will improve the flash talk. \newline 2. Preserve essential placeholders in the updated prompt, ensuring that necessary variables, which are wrapped with curly brackets (\{ \}), remain intact. \newline 3. Adapt the structure of the flash talk dynamically based on the feedback. Instead of always using 'Introduction, Methodology, Significance, and Conclusion', design sections that best reflect the improvements needed. \newline 4. Maintain an engaging and accessible tone, ensuring the output remains clear, concise, and compelling for a public audience. \newline 5. Ensure the revised prompt is formatted with two key components: system\_prompt and user\_prompt. \\
\midrule

Evaluation & Evaluation \newline Agent & Assess \newline video clip \newline quality & Evaluation Scores & \# Content Accuracy Evaluation Example: \newline You are an expert evaluator of scientific content accuracy. Your task is to assess whether the video accurately represents the key aspects of a research paper without distortion or misinterpretation. \newline You will analyze the scientific integrity and conceptual coverage of the video based on the provided research paper TEXT summary and video frames.

Audio Script: \texttt{<\textbar audio\_script\textbar>}

Evaluate the content accuracy of this video with the audio\_script based on the following aspects:
\newline - Scientific Integrity: Does the video accurately represent the research findings without distortion or misinterpretation?
\newline - Key Concept Coverage: Does it cover the core aspects of the research (hypotheses, methodology, key results, and conclusions) effectively?

Here is the paper summary of the scientific paper this video is describing \texttt{\{paper\_summary\}}

Rate BOTH aspects on a scale of 1 to 5 (1 = Poor, 2 = Acceptable, 3 = Good, 4 = Great, 5 = Excellent). Do not add your explanation. Just give your scores on each aspect.

[Output Format]
\newline Scientific Integrity: [SCORE]
\newline Key Concept Coverage: [SCORE]
\\
\bottomrule

\caption{Comparison of different agents.}
\label{tab:agent_info}
\end{longtable}
\end{tiny}

\begin{tiny}
\begin{longtable}{p{1.1cm}|p{1.1cm}|p{1.16cm}|p{9cm}}
\toprule
\textbf{Type} & \textbf{Evaluation Section} & \textbf{Metric} & \textbf{Description} \\ \midrule
\midrule
\endhead
Feedback & Flashtalk & Clarity & - Is the message clear and easy to follow? \\
\cline{3-4}

 &  &  & \\
 &  & Curiosity & - Does the hook immediately capture audience interest? \\
\cline{3-4}
 
 &  &  & \\
 &  & Effectiveness & - Does the flash talk motivate viewers to explore the full content? \\
\cmidrule{2-4}

 & Sceneplan & Narrative Coherence & - Does the sequence of sub-scenes follow a logical flow?
- Are transitions smooth, avoiding abrupt shifts between scenes?
- Does the structure maintain audience engagement throughout the video? \\
\cline{3-4}
 
 &  &  & \\
 &  & Timing and Pacing & - Are the durations of sub-scenes appropriately distributed?
- Do any sections feel rushed or overly extended?
- Is the pacing consistent, maintaining viewer engagement? \\
\cline{3-4}

 &  &  & \\
 &  & Visual Relevance and Clarity & - Do the selected images and visuals align with the sub-scene descriptions?
- Are there any unclear or ambiguous visual choices?
- Does each sub-scene effectively illustrate the corresponding narration? \\
\cmidrule{2-4}

 & Text & Clarity & - Are the text components clear, concise, and logically structured?
- Do they enhance scene understanding without clutter? \\
\cline{3-4}

 &  &  & \\
 &  & Key Information Coverage & - Do the extracted texts effectively summarize core ideas from the scene? \\
\cline{3-4}

 &  &  & \\
 &  & Timing and Alignment & - Are text components timed appropriately with the audio narration and visuals?
- Do the texts appear and disappear naturally to support visual storytelling? \\
\bottomrule

\caption{Evaluation rubrics for \textit{Feedback Agents}.}
\label{tab:eval_rubric_feedback}
\end{longtable}
\end{tiny}

\begin{algorithm}[t]
\caption{Feedback Aggregation and Updating Agent Prompts}
\label{alg:feedback-loop}
\begin{algorithmic}[1]
\Require Initial Prompt $P^{(0)}$, Video $\mathcal{V}$, Feedback Agents $\mathcal{A} = \{A_{Flashtalk}, A_{ScenePlan}, A_{Text}\}$, Flashtalk, Sceneplan, and ReflectionFunction
\Ensure Optimized Prompt for next iteration
\For{iteration $n = 1, 2, \dots, N$}
    \State feedback\_list $\gets$ empty
    \For{scene in Sceneplan}
        \For{sub\_scene in scene.sub\_scenes}
            \For{agent in [Flashtalk, Sceneplan, Text]}
                \State feedback\_metrics $\gets$ SetMetrics(sub\_scene)
                \State feedback $\gets$ PromptFeedbackAgent(agent, video(sub\_scene), metrics)
                \State Append feedback to feedback\_list
            \EndFor
        \EndFor
    \EndFor
    \State summarized\_feedback $\gets$ Summarize(feedback\_list)
    \For{agent in agents}
        \State agent.prompt $\gets$ ReflectionFunction(agent.prompt, summarized\_feedback)
    \EndFor
\EndFor
\State \Return updated agent prompts
\end{algorithmic}
\end{algorithm}

\end{document}